\pdfoutput=1

\documentclass[11pt]{article}

\usepackage{naacl2021}

\usepackage{times}
\usepackage{latexsym}
\usepackage{graphicx} 
\usepackage{svg}
\usepackage{subcaption}
\usepackage{rotating}
\usepackage{hyperref}
\usepackage{pdfpages}

\usepackage[T1]{fontenc}

\usepackage[utf8]{inputenc}

\usepackage{microtype}

\usepackage{inconsolata}

\title{Graph Integrated Language Transformers for Next Action Prediction in Complex Phone Calls}

\author{Amin Hosseiny Marani {\normalfont and} Ulrike Schnaithmann {\normalfont and} Youngseo Son {\normalfont and}\\
\textbf{Akil Iyer} and \textbf{Manas Paldhe} and \textbf{Arushi Raghuvanshi} \\
  Infinitus Systems, Inc. \\
\texttt{ \{amin.hosseiny, ulie.schnaithmann, youngseo.son, akil.iyer,} \\
\texttt{ manas.paldhe, arushi\}@infinitus.ai}\\
  }

\begin{document}
\maketitle
\begin{abstract}

Current Conversational AI systems employ different machine learning pipelines, as well as external knowledge sources and business logic to predict the next action. Maintaining various components in dialogue managers' pipeline adds complexity in expansion and updates, increases processing time, and causes additive noise through the pipeline that can lead to incorrect \textit{next action prediction}. 
This paper investigates graph integration into language transformers to improve understanding the relationships between humans' utterances, previous, and next actions without the dependency on external sources or components.

Experimental analyses on real calls indicate that the proposed Graph Integrated Language Transformer models can achieve higher performance compared to other production level conversational AI systems in driving interactive calls with human users in real-world settings.

\end{abstract}

\section{Introduction} \label{sec:intro}

\begin{figure}
    \centering
    \includegraphics[width=1.0\linewidth]{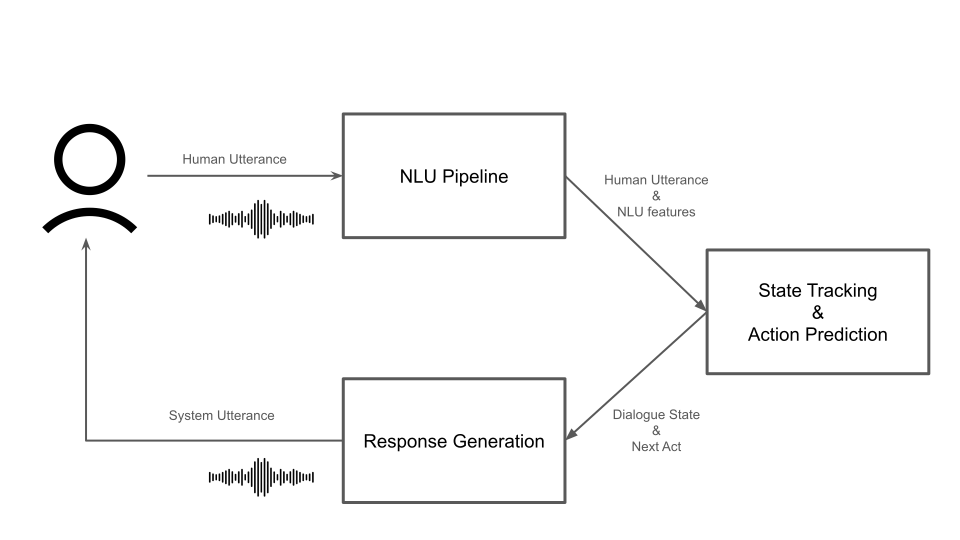}
    \caption{A schematic visualization of dialogue managers' components which utilize an NLU pipeline of models to extract intents and fill slots from human utterances, and predict the next action based on the current and previous state. Finally, the system generates an utterance to respond to the human users (e.g., using LLMs or predefined templates).}
    \label{fig:dialogue_manager}
\end{figure}

Building and maintaining complex production quality conversational systems has been an ongoing challenge in industry. One approach to solve complex conversational tasks such as outbound call automation, is to use a dialogue manager~\cite{paek2008automating, teixeira2021interplay} to encode business logic.
Conversational systems which use dialogue managers have multiple components which consist of Natural Language Understanding (NLU)~\cite{bocklisch2017rasa}, dialogue state tracking~\cite{mannekote2023towards}, next action prediction~\cite{mannekote2023towards}, and response generation~\cite{weston-etal-2022-generative,he-etal-2018-decoupling}. Figure~\ref{fig:dialogue_manager} describes the process of call automation systems with the aforementioned components.

\begin{figure}
    \centering
    \includegraphics[width=1.2\linewidth]{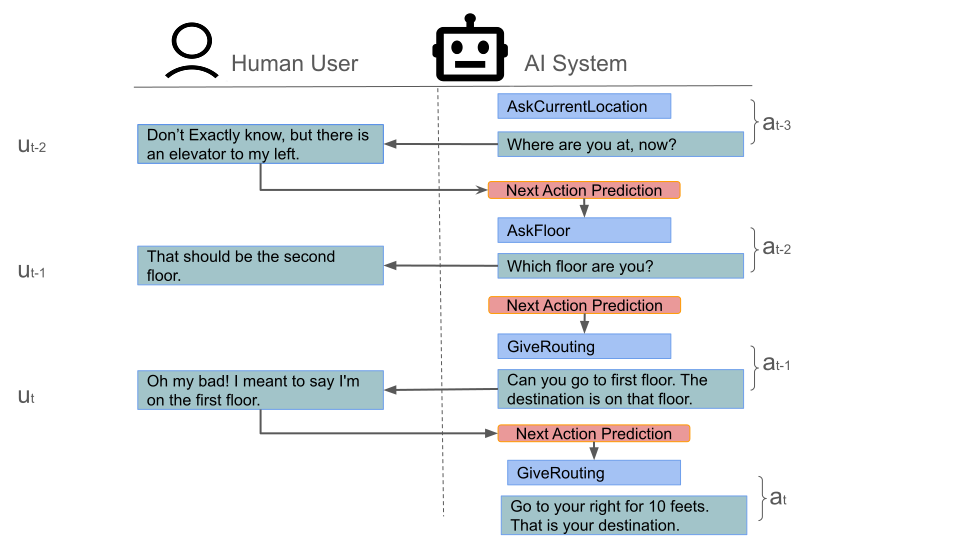}
    \caption{An example of Visual Navigation Assistant as a dialogue manager. At each time-step $t$,
     the dialogue manager extracts the entities such as slots and intents from human utterance $u_t$ (i.e., {\color{teal} green rectangles} on the left side) and predicts the next action $a_t$ (i.e., {\color{red} red rectangles}). Using the predicted action (i.e., {\color{cyan} blue rectangles}) the dialogue manager generates a system response (i.e., {\color{teal} green rectangles} on the right side).} 
    \label{fig:example}
\end{figure}

Handling \textit{next action prediction} is one of the critical tasks dialogue managers take care of, as it affects the response generation directly~\cite{traum2017computational_survey}. 
\textit{Next action prediction} is the process of analyzing human utterance, current and previous state of the conversation (i.e., dialogue state tracking) and deciding which action to take, which in many industry settings is returning a specific response template. 
Figure~\ref{fig:example} demonstrates an example of a dialogue manager based conversation automation as a visual navigation assistant for multiple dialogue turns.

Recently, there has been significant progress in the field of Generative AI and Large Language Models (LLMs) for end-to-end conversational systems which alleviate the need for manually engineered dialogue managers~\cite{mannekote2023towards,snell-etal-2022-context}. However, they sometimes have issues with hallucinations and can underperform in domain specific, targeted conversations such as those that require knowledge graph retrieval~\cite{dziri2021neural,ji2023survey}.

In most industry settings, templates are used with action prediction to generate the response.
By predicting an action, we are determining which response template(s) to return to the user~\cite{mannekote2023towards,qiu2022towards_template_action,urbanek-etal-2019-learning_template_action}. Action prediction using response templates instead of language generation helps prevent hallucinations, adds necessary guardrails for some industry settings, and keeps latency low.

To solve the \textit{next action prediction} problem, different NLP methods from traditional symbolic AI techniques such as Knowledge Graph models~\cite{he2017learning_knowledge_graph,de2018talk_symbolic_ai}, to more modern transformer based techniques~\cite[e.g.,][]{zhou2023cast_dialogue_act_llm} have been introduced; however, two main challenges still persist.
1) a majority of prior work depends on Slot-Filling (SF) and Intent-Classification (IC) techniques to extract dependencies and relies on external sources (i.e., knowledge or rule based approaches) to find the relationship between the extracted information and actions~\cite{mannekote2023towards,traum2017computational_survey}. Instability in detecting SF and IC causes incorrect \textit{next action prediction}.
2) many conversational systems handle grounding poorly~\cite{traum2017computational_survey,weston2018retrieve_dialogue_managers_challenge,sutskever2014sequence_dialogue_managers_challenge}; this is when users' responses differ from expected inputs (e.g., referring to a previous point in the conversation, moving backwards to change a previous response, or no action-related slots being detected).
For example, in Figure~\ref{fig:example}, the human user sets a new ground by mentioning the elevator instead of the their location. This information may be slightly different than what a \textit{next action prediction} model expects and can respond to.
Lack of grounding in a conversation and more specifically in a model may result in misunderstanding~\cite{traum2017computational_survey} and can damage the conversation.

This paper introduces an approach to predict the next action without any dependency on information extraction (i.e., SF and IC) or external resources~\footnote{The proposed model is trained using external resources but does not need any external resources after training.} such as ontology~\citep[e.g.,][]{altinok2018ontology} or knowledge-base~\citep[e.g.,][]{vizcarra2022knowledge} approaches. 
The proposed models, Graph Integrated Language Transformers, learns co-occurrences of actions and human utterances through a graph component (i.e., Graph Neural Network or a graph embedding layer) and combines it with language transformers to add language understanding in production settings.
The model is trained on conversations that followed a Standard Operating Procedure (SOP)~\footnote{SOP is a document which defines a set of guideline instructions for diverse situations during the conversations.} without the need for explicit encoding.
The proposed model can be trained on any similar dataset that has an inherent action-to-action relationships.
The list below summarizes the contribution of this paper.

\begin{itemize}
    
    \item Integrating graph information and combining with language transformers to remove dependency on NLU pipelines.

    \item Adding a graph component (i.e., history of action co-occurrence) to language transformers to predict the next action as one atomic task while also overcoming the token limit by removing the need to keep prior dialogue history. 

    \item Evaluating the proposed \textit{next action prediction} model in a production setting against a system that relies on an NLU pipeline with an explicitly defined dialogue manager (DM system) in Appendix~\ref{app:common_system}.

\end{itemize}

To examine the performance and robustness of the proposed models in real-world settings with noisy input, the evaluation is done in a production setting and goes beyond classification metrics; the evaluation includes industry critical factors such as human experience using the conversational system and considers real-time constraints such as latency of output generation.

\section{Related Work} \label{sec:related_work}

\textit{Next action prediction} approaches can be categorized in three chief groups. 
First, structured-based approaches that consider sequential relationships between previous actions, other actions, and their requirements. 
These approaches assume that the current state (i.e., the previous action) is known~\cite{henderson2015discriminative}.
On the one hand, local structure-based approaches such as \textit{Question \& Answer} systems~\cite{reshmi2016implementation} consider local adjacency of the actions, utterance features, and next potential actions. 

On the other hand, global structured-based approaches define problem space using dialogue-grammars or finite-state networks~\cite{traum2017computational_survey,wollny2021we}. 
However, none of structured-based approaches provide the ability to train a model and they require expert to design them~\cite{henderson2015discriminative}.

The second group of \textit{next action prediction} approaches are principle-based.
These techniques choose next actions based on the filled information rather than sequential order between actions, thus behaving both locally and globally~\cite{traum2017computational_survey}. 

Slot-filling (SF) and Intent-classification (IC) based techniques (i.e., joined or separate components) are common principle based approaches~\cite{louvan2020recent}.

Recently, neural models including RNNs and Language Transformers which act solely on input are receiving more attention for SF-IC based techniques~\cite{goo2018slot,chen2019bert,zhang-wang-2022-srcb}. These methods are mainly using dialogue history alongside additional information such as schema of the task (e.g., ``hotel booking'' or ``scheduling a doctor's appointment'') e.g., using embedding layers with or without attention layers fused with a language transformer~\cite[e.g.,][]{mosig2020star,mehri2021schema, zhang2021sgd}. 

However, most of these language transformer based techniques were only evaluated on datasets with low number of actions, 10 or less~\cite{mosig2020star,rastogi2020sgd}, or perform poorly on larger number of actions (i.e., 30 actions) for one top output selection~\cite{chen2021action}.

\section{Methodology} \label{sec:method}

This section discusses the problem definition of the \textit{next action prediction} task (i.e., Section~\ref{sec:methods_problem}), 
and introduces the proposed models (i.e., Section~
\ref{sec:methods_GaLT}).

\subsection{Problem Definition}\label{sec:methods_problem}

A \textit{next action prediction} model chooses an action $a_t$ given $U_{k:t}$ and $Z_{k:t-1}$ at time $t$ in which $U$ is the set of all utterances from time $k$ (i.e., $k\geq0$) to time $t$, and $Z$ is the set of all previously predicted acts. 
Equation~\ref{eq:1} formulates the process of \textit{next action prediction}. In this equation, $f$ denotes any function (e.g., machine learning model or a probabilistic matching technique) that can map thereof inputs to the next action.

\begin{equation} \label{eq:1}
    a_{t} = f([U_{k:t},Z_{k:t-1}]) 
    ~~~s.t. ~~~ 0\leq k \leq t-1 
\end{equation}

Different techniques approach \textit{next action prediction} differently. Some techniques rely on feature extraction from utterances (i.e., $U_{k:t}$) using NLU techniques (e.g., intents or slots in NLU pipeline of Figure~\ref{fig:dialogue_manager}); in those cases $U_t$ in Equation~\ref{eq:1} becomes utterance and all those extracted features at time $t$.
However, this paper proposes a method that relies only on the very last human utterance 
and previous actions in Section~\ref{sec:methods_GaLT}.


\subsection{Graph Integrated Language Transformers} \label{sec:methods_GaLT}

This paper proposes a graph integrated approach to employ the rich information of graph-like structures, discussed in Section~\ref{sec:related_work} (e.g., SOP, graphs, or rule knowledge bases) and combine it with language transformers. 
Two different techniques are proposed in this section that each combine language transformers with 1) Graph Neural Networks (GNN) to explicitly encode the graph of actions and other features (GNN-LT), and 2) a graph embedding layer to learn co-occurrences of action history, Graph-aware Language Transformer (GaLT).

Both models additionally use language transformers such as BERT~\cite{devlin2018bert}, DistilBERT~\cite{sanh2019distilbert}, or RoBERTa~\cite{liu2019roberta} to add language understanding~\cite{devlin2018bert} to the \textit{next action prediction}.
The GNN-LT models is fed past actions as nodes and features of nodes' connections as edges (i.e., order of the connections, slots, and embedding of the utterance) using a Graph Attention Network~\cite{yun2019graphormer}. Thus, GNN-LT explicitly integrates the graph knowledge including the order of the actions and their connections.
GaLT employs a graph embedding layer that encodes past actions as node labels directly without the past action names or utterances; therefore implicitly adds the ability to learn the co-occurring utterances and actions without the need to explicitly enforce graph constraints (i.e., actions as nodes, filled slots or other features as edges).
Additionally, GaLT acquires fewer training parameters (e.g., $66M$ Distilbert + $1M$ fusion and fully connected layer = $67M$ in total) in comparison to GNN-LT (e.g., $66M$ Distilbert + $12M$ Graphormer small~\cite{yun2019graphormer} +  $1M$ fusion and fully connected layer = $79M$ in total); therefore, GaLT requires less training time and performs much faster in inference.

The language transformer is fed the human utterance alongside the history of actions to implicitly learn the co-occurrence between human responses and follow-up actions taken by the system. 
Additionally, the language transformer is pre-trained on a much larger dataset of full dialogue turns to learn the context of the utterances and their co-occurring actions. 
As the dialogue history is removed from the graph integrated language transformer training process, the model is incentivised to focus on action co-occurrence and sequences as graph nodes rather than the dialogue history surrounding them. 

\begin{figure*}
    \centering
    \begin{subfigure}[b]{0.45\textwidth}
        \centering
        \includegraphics[width=7cm]{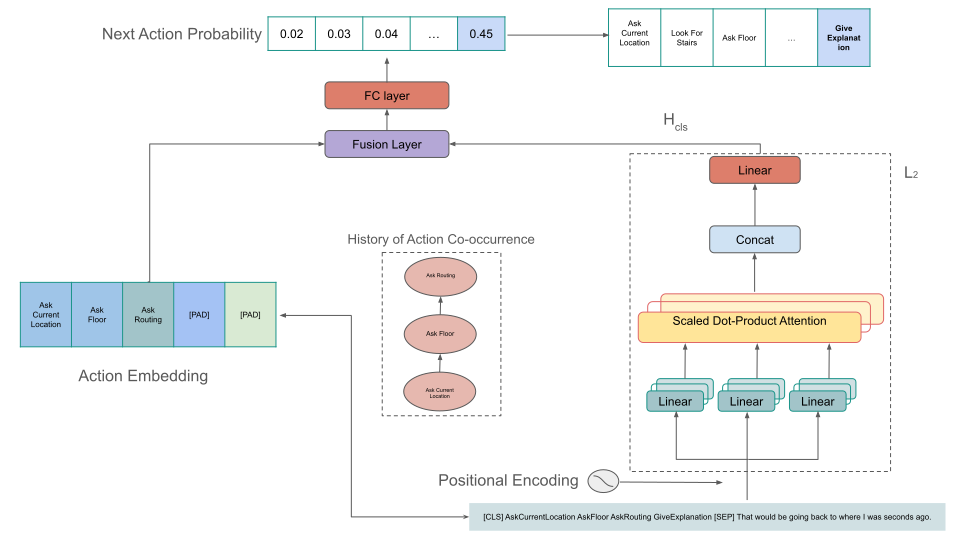}
        \label{fig:current_panel_num}
    \end{subfigure}
    \begin{subfigure}[b]{0.45\textwidth}
        \centering
        \includegraphics[width=7cm]{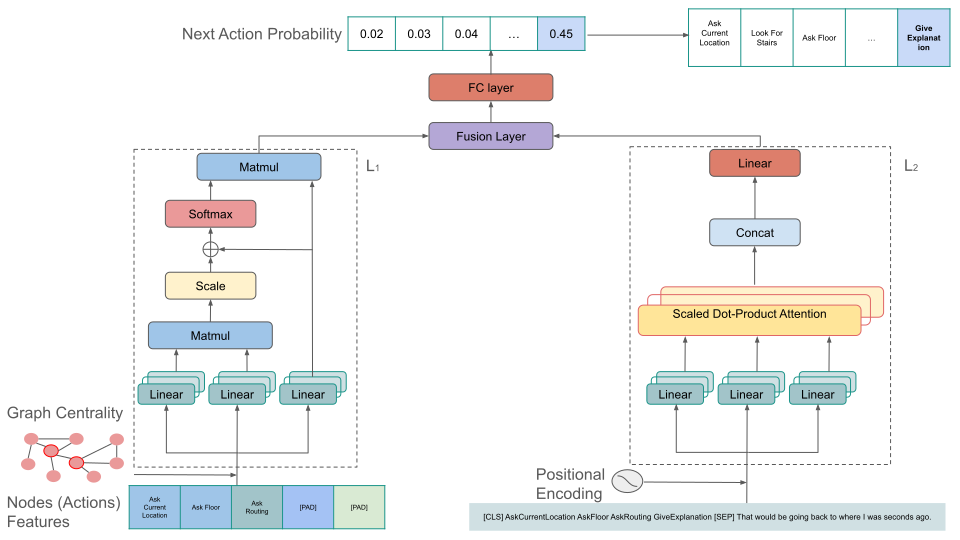}
        \label{fig:proposed_panel}
    \end{subfigure}
    \caption{The architecture of the GaLT model (i.e., left figure) and GNN-LT (i.e., right figure). GaLT is fed action history as graph embedding and GNN-LT is fed actions as nodes as well as utterance features as edges; each models then is fused with a language transformer. $L1$ denotes the number of layers in GNN and $L_2$ denotes the layers of the language transformer.}
    \label{fig:gnn-llm}
\end{figure*}

Keeping only actions as the history of the dialogues (i.e., both in language transformer and graph components) removes dependency to the NLU pipeline (i.e., discussed in Section~\ref{sec:intro} and \ref{sec:related_work}) and the need to keep the dialogue turns' utterances; thus improving speed of prediction and satisfying the language transformer token limit; e.g., 512 for DistilBERT~\cite{sanh2019distilbert,devlin2018bert}.
Due to the simplicity of the model, real time inference time requirements are still being met. 
Figure~\ref{fig:gnn-llm} shows a schematic of the proposed models.

A fusion layer combines both language transformer and graph component features using Equations~\ref{eq:lt_hidden}-\ref{eq:fusion}. 
First, Equation~\ref{eq:lt_hidden} computes mean of the hidden features from the language transformer and Equation~\ref{eq:embedding_hidden} computes the features of the graph component. 
Here, $W$ and $b$ are trainable parameters, $O$ is the output of a layer, $l$ and $g$ denote the language transformer and graph component.
Then, the fused features will be fed into a fully connected layer to predict the next action. 
Equation~\ref{eq:fusion} fuses the hidden features of both layers and generates the probability using the $Softmax$ activation layer.
The next action will be picked from the list of all actions with respect to their probability of the computed Softmax output.
While there are variety of fusion techniques (e.g., concatenation, dot product techniques, or summation techniques), Equation~\ref{eq:fusion} uses $\otimes$; since GaLT and GNN-LT reach to the highest performance via pairwise dot product fusuion.

{\small
\setlength{\abovedisplayskip}{0pt}
\setlength{\belowdisplayskip}{0pt}

\begin{equation}\label{eq:lt_hidden}
    H_l = GELU(W_l ~~ mean(O_l) + b_l)
\end{equation}

\begin{equation}\label{eq:embedding_hidden}
    H_g = GELU(W_g O_g + b_g)
\end{equation}

\begin{equation}\label{eq:fusion}
    H_f = Softmax(W_f (H_l \otimes H_g) + b_f)
\end{equation}
}

\section{Experimental Setup and Results} \label{sec:results}

This section describes the process of collecting data for training the models, comparing the trained models regarding classification metrics (i.e., $F1$), and evaluating the proposed models as well as the DM system\footnote{The current production system that is handling the call automation at the time is called DM system throughout this paper.}, explained in detail in Appendix~\ref{app:common_system}, using a human-centered approach.

\subsection{Data, Configurations, and Training} \label{sec:results_data_config}

To integrate the graph information into GNN-LT and GaLT models, this work utilizes conversational data which follows a Standard Operating Procedure (SOP). These conversations were guided by a human expert or the DM system which employs a human defined SOP. The SOP is a graph like structure with actions as nodes and their connections to next actions based on filled slots,  which has been carefully translated into dialogue manager logic. 
Appendix~\ref{app:sop} discuss the SOP in more details.
However, the proposed Graph Integrated Language Transformers were not trained on the SOP explicitly. GaLT and GNN-LT were trained on the data human experts and the DM system collected and generated from the SOP.

To evaluate the proposed models, dialogue turns of phone calls between human-AI and human-human were collected from June to August 2023.

The next action for each human dialogue turn was decided and labeled by the DM system with human in the loop supervision. Human domain experts intervened in calls that might fail. The intervention varied from correcting the collected data (e.g., spelling mistakes) to driving the calls in severe cases.
To generate a reliable dataset, a team of human experts classified each conversation as successful or unsuccessful on a call level, rather than labeling and reviewing each dialogue turn, due to financial reasons and limited human resources. 
For the same reason, all of dialogue turns for each call are added to the dataset if it was considered successful~\footnote{If the model managed to prompt the human user to give all information required} or was dropped otherwise.
That resulted in $\sim1M$ records each including one human utterance and one system response.
In addition to selecting successful calls, a pre-processing step (described in Appendix~\ref{app:preprocessing}), is devised to remove undesirable dialogue turns, calls, or actions; e.g., actions that are deprecated and the rest of the call to avoid incorrect connection between actions.
This process led to $\sim600K$ remaining dialogue turns. 

Despite filtering out $\sim400k$ dialogue turns, the language transformers were initially pre-trained on all dialogue turns (i.e., $\sim1M$) using Masked Language Modeling (MLM)~\cite{devlin2018bert} and then fine-tuned on $\sim600K$ selected dataset for the \textit{next action prediction} task.
The dataset was randomly split to 80\%-10\%-10\% for training, validation, and test. 
Section~\ref{app:datset} summarizes the details of the dataset. Additionally, Section~\ref{app:sys_config} and Section~\ref{app:other_model_params} lists the system configurations and proposed models' hyper-parameters for training and testing the models.

\subsection{Classification Performance Comparison} \label{sec:results_metrics_evaluation}

This section evaluates the proposed models and other techniques using an offline classification evaluation. The process evaluates each technique's performance on the turn-level; next action given a human user's utterance and the previous actions or dialogue history. 
To measure the performance for each model, F1 Score was computed on the test-set described in Section~\ref{sec:results_data_config}.

Table~\ref{tab:f1} compares the proposed models with other techniques.
The dataset, described in Appendix~\ref{app:datset}, consists of 80 next actions (i.e., classes) of imbalanced frequency; thus $F1_{Macro}$ was calculated alongside $F1_{weighted}$. 
The results suggest that stand-alone models (i.e., language transformers or GNNs) and prompt-based large language models~\footnote{This paper also evaluates a prompting only approach using Llama2 (\href{https://ai.meta.com/llama}{https://ai.meta.com/llama}) on the same task and dataset; however, the results are not reported due to poor results in comparisons with other models. The prompt that is used to generate outputs as well as the results are discussed in Appendix~\ref{app:prompt}.} are not able to predict the next action with high performance (i.e., lower $F1_{macro}$). 
Moreover, this table shows adding the graph embedding of actions in GaLT can improve $F1$ for \textit{next action prediction} more than combining complex GNN models.
GaLT also can reach to its high performance with as little as $60K$ dialogue turns(i.e., 10\% data size) as described in Appendix~\ref{app:data_effect}.

\subsection{Human-Centered Evaluation} \label{sec:results_human_centered_evaluation}

\begin{table}
    \small
    \centering
    \begin{tabular}{p{4cm}|ll}
         Model & \tiny $F1_{Weighted}$ & \tiny $F1_{Macro}$\\
         \hline
         BERT w/ dialogue history~\cite{mosig2020star} & 0.58 & 0.38 \\
         BERT w/ SF~\cite{zhang2021sgd}  & 0.79 & 0.44 \\
         \hline
         BERT w/ action history & 0.80 & 0.63 \\
         DistilBERT w/ action history & 0.82 & 0.69 \\
         RoBERTa w/ action history & 0.78 & 0.60 \\
         \hline
         GNN~\cite{yun2019graphormer} & 0.72 & 0.52 \\
         (sub)\textsuperscript{*}GNN~\cite{yun2019graphormer}  & 0.72 & 0.51 \\
         \hline
         GNN-LT(DistilBERT) & \underline{0.84} & 0.72 \\
         (sub)\textsuperscript{*}GNN-LT(DistilBERT) & \underline{0.84} & 0.72 \\
         GaLT & \underline{0.84} &\underline{0.75} \\
         \hline
         \multicolumn{3}{l}{*sub-GNN models are fed only recent actions} \\
         \multicolumn{3}{l}{~~~(e.g., last 5 or 10).}
    \end{tabular}
    \caption{Summary of offline classification evaluation across different techniques regarding $F1$. Four categories of models were listed in this table; language transformers (e.g., BERT) with dialogue history or detected filled slots, language transformers with last utterance and recent history of actions (e.g., 5 or 10 last actions), GNN model, and Graph Integrated language transformers (e.g., GNN or graph embedding). The \underline{underscore} values show the best performance regarding each metric (i.e., columns).}
    \label{tab:f1}
\end{table}

This section evaluates the best performing model, GaLT, with the DM system using a human-centered approach since the desired outcome of a call can be achieved through various paths and does not need to be strictly tied to one correct next action (i.e., what was done in Section~\ref{sec:results_metrics_evaluation}).
Put precisely, more than one next action can be considered as a correct prediction given the recent actions and the current utterance.
To compare GaLT with the DM system, human assessors acted the ``role'' of the agent receiving outbound calls. They were familiar with the call structure and expected outcome of calls.
Two different approaches were designed to compare and evaluate the models; objective product-level and human subjective.
Additionally to test the generalizability and robustness of the compared models three call difficulty levels were defined; easy, medium, and hard (i.e., Table~\ref{tab:difficulty-levels}). As the call difficulty level increases human utterances and provided information get more complex (e.g., mumbling or updating a piece of information).
The experimental setup and metrics are described in more detail in Appendix~\ref{app:human_centered_evaluation_setup}.

\paragraph{Production Level Metrics}

Table~\ref{tab:product_metrics} shows that the proposed models outperforms the DM system regarding both \textit{field number} (i.e., how much information the call collected) and \textit{panel number~\footnote{Panel number indicates the progress a model is made into finishing a call. Panel 0,1,2,3,4, and E2E indicate 0\%, 20\%, 40\%, 60\%, 80\%, and 100\% progress of a call respectively.}} (i.e., how far to the end of the call model reached).
T-test statistics analysis suggests that the comparisons were significant for \textit{medium} level as well as all levels combined (i.e., `.' and `*' symbols for each pair in Table~\ref{tab:product_metrics}).
In addition to the \textit{panel number}, finishing a call successfully (e.g., collecting all information or without human user hanging up) is another important metric (i.e., \textit{E2E} metric). GaLT also improved the \textit{E2E} or number of successfully finished calls by +31.92\% (Appendix~\ref{app:extended_product_results} shows an extensive comparison).

\begin{table}
    \centering
    \small
    \begin{tabular}{p{1cm}|p{1.15cm} p{1.30cm}|p{1.15cm} p{1.2cm}}
         Difficulty & \multicolumn{2}{c|}{\#Fields Mean (std)} & \multicolumn{2}{c}{\#Panels Mean(std)} \\
         ~~Level & DM system & Proposed & DM system & Proposed \\
         \hline
         Easy & 23.1(6.59) & 25.35(6.19) & 3.85(0.65) & 4.0(0.0)\\
         Medium & 18.36(9.23)\textsuperscript{.} & 23.3(4.45)\textsuperscript{.} & 3.05(1.39)\textsuperscript{**} &(4.0)0.0\textsuperscript{**}\\
         Hard & 18.25(5.49) & 21.44(5.98) & 3.63(0.99) & 3.66(0.94) \\
         \hline
         Total & 20.36(7.97)\textsuperscript{*} & 23.79(5.70)\textsuperscript{*} & 3.48(1.13)\textsuperscript{*} &3.93(0.42)\textsuperscript{*}\\
         \hline
         \multicolumn{5}{l}{\small \textit{Note: 
         \textsuperscript{.}p<0.1, \textsuperscript{*}p<0.05, \textsuperscript{**}p<0.01, \textsuperscript{***}P<0.001}}
    \end{tabular}
    \caption{Comparing the DM system and proposed models performance regarding product-level metrics, number of fields and panels, across different difficulty levels. The results of t-test are shown as stars (\textit{`*'}) or dots (\textit{`.'})}
    \label{tab:product_metrics}
\end{table}

\paragraph{Subjective Human Evaluation} 

Additionally, Human agents (i.e., human users who interacted with the models) and reviewers rated each call after finishing that call as described in Appendix~\ref{app:human_centered_evaluation_setup} using a 5-point Likert scale rating.
The GaLT model received a higher rating average of 2.91 ($std=1.15$) in comparison to rating average of 2.78 ($std=1.42$) for the DM system.
Comparing the number of positive and negative ratings for each model shows that both models received almost same number of positive ratings but the DM system received higher number of negative ratings. In other words, human assessors rated the proposed models to be more robust. A deeper investigation regarding difficulty levels is done and discussed in Section~\ref{app:extended_subjective_results}.

\section{Conclusion} \label{sec:conclusion}
\label{sec:bibtex}

This paper proposes Graph Integrated Language Transformers technique to improve \textit{next action prediction} performance
to resolve the dependency 
on Slot-Filling and Intent-Classification techniques 
and grounding issue~\cite{mannekote2023towards}.
The analyses indicate that keeping the action history with order of the actions using a graph embedding layer and combining with language transformers generates higher quality of outputs in comparison to more complex models that include connection details of actions (i.e, GNNs including the connection details through edges).
The proposed model(s) improve the \textit{next action prediction} regarding $F1$ as well as product-level metrics and human-centered evaluation.
They can improve the robustness regarding \textit{next action prediction} (e.g., less unexpected results or being stuck in a loop) in comparison to other techniques and handle complex tasks better in comparison to the DM system in long noisy phone calls.
Additionally, the proposed models can reach to a high performance level with as low as $60K$ dialogue turns.
We hope future research can employ a similar method combined with generative AI models to extract the information from human utterances as well as generating custom responses to automate calls without dependency on other components.

\section*{Limitations}

Although the proposed models can reach to a high performance with as little as $60K$ dialogue turns, it needs re-training or fine-tuning for any new application in a new domain or even with slightest changes; e.g., adding or removing even one action.
Moreover, similar to other neural models, graph integrated language transformers, lack interpretability and may show instability (e.g., predict an action that does not have any relationship to dialogue history).

In addition to these limitations, the evaluation can benefit from further investigation.
This paper recruits human agents and employees who were familiar with the DM system. That can lead to a biased assessment and perhaps is the source of inconsistency between human subjective rating and product-level metrics. 

Finally, there are next steps to further evaluate graph injection with additional third party GenAI prompt based models. The ability to use certain third party systems was limited at the time of evaluation due to the requirement for this healthcare dataset to stay HIPAA compliant.

\bibliography{anthology,custom}
\bibliographystyle{acl_natbib}

\appendix

\section{Conversational Systems' \textit{Next Action Prediction} Process}\label{app:common_system}

\begin{figure*}
    \centering
    \includegraphics[width=1.0\linewidth]{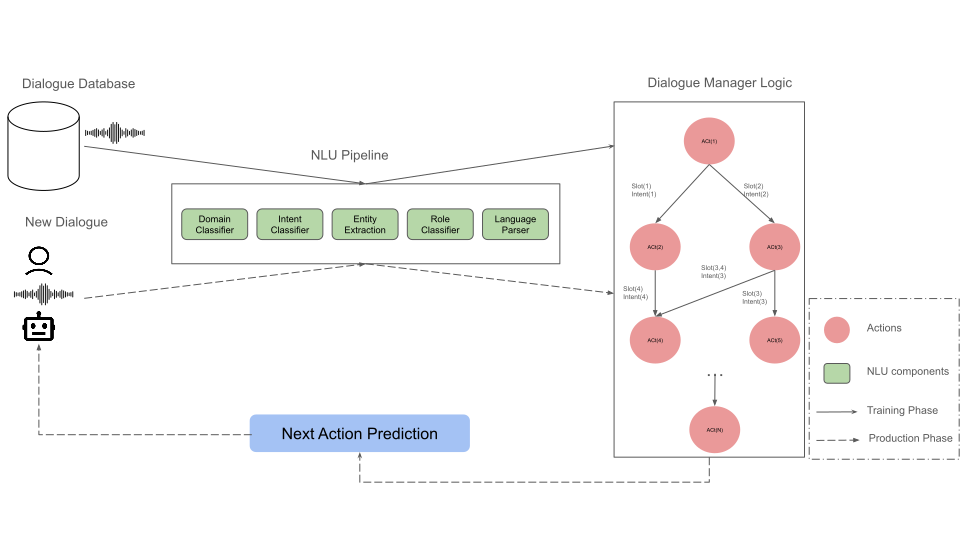}
    \caption{Overview of \textit{next action prediction} process in a conversational AI system. 
    The dialogue manager logic is generated using previous dialogues and via NLU pipeline (e.g., SF and IC). The model predicts next action using incoming utterances, NLU pipeline, and the generated knowledge.
    The arrows in the image show the connections between data, NLU pipeline, and dialogue manager logic during training (i.e., straight lines) or prediction during production (dotted lines). 
    }
    \label{fig:common_system}
\end{figure*}

Figure~\ref{fig:common_system} shows a schematic overview of how a \textit{next action prediction} model is employed in conversational AI systems. 
Although different conversational AI systems may use different approaches for NLU analysis or dialogue manager logic, most of the current approaches still use similar mechanism~\cite[e.g.,][]{mannekote2023towards, bocklisch2017rasa, raghuvanshi2018developing_mindmeld}.


\section{Standard Operating Procedure}\label{app:sop}

The DM system described in Appendix~\ref{app:common_system} employs a human defined Standard Operating Procedure (SOP) to guide the conversation based on the last action and conversational context of past slots filled. For example, one of the questions asked by the AI system is "Is this a commercial or government plan?" Depending on the type of plan different paths have to be followed. If it is a government plan, the AI system should ask "Is this Medicare, Medicaid, or Tricare?". If it is a commercial plan, the AI system should next ask about the Rx Number. If the Rx number is the same as a previously provided policy number, the AI system should push back to clarify "Just to confirm, the RX group number and the policy number are the same?". Similarly, throughout the conversation these types of guidelines are defined which are necessary for collecting accurate information in these healthcare calls. Depending on the information provided so far on the call, the SOP may require different confirmations and followup loops similar to the above example.

\section{Data Preprocessing}
\label{app:preprocessing}

The data preprocessing resulted in $>593K$ records each including one human utterance, the previous and next action as well as the system response.
Through preprocessing, three types of records were removed from the dataset:
\begin{itemize}
    \item records with rare or obsolete~\footnote{No longer has been used in the DM system} next actions and the rest of the call: A low number of next actions, $N~10$, only appeared less than 50 times across  the dataset due to different reasons (e.g., getting merged or updated). While the preprocessing kept the dialogue history up to that moment, the rest of dialogue was dropped since lack of prior information (i.e., deleted records) can be misleading for a \textit{next action prediction} model. 

    \item records with filler actions such as \textit{wait, just a moment, or repeat last sentence} : 
    The preprocessing also dropped these records and actions because, filler actions 1) do not add any meaningful instructions to the graph structure and 2) do not need dialogue history or previous actions to be detected.

   The preprocessing also dropped these for the same reasons stated for \textit{waiting} actions above. 
\end{itemize}

In addition to these steps, utterances split into fragments (i.e., multiple dialogue turns with one same next action) were merged to form one record with one desired next action.

Although, it is important to handle edge cases such as incomplete sentences for a conversational AI system in call automation, managing those are less relevant to the \textit{next action prediction} models.
Moreover, the proposed models handled incomplete sentences  well during evaluation.

\section{Dataset Details} \label{app:datset}

This section summarizes the details of the dataset regarding number of calls and dialogue turns in Table~\ref{tab:dialogue_info} as well as actions and panels in Table~\ref{tab:action_info}.

\begin{table}
    \small
    \centering
    \begin{tabular}
    {ll}
    \hline
    Calls & 21,220 \\
    \hline
    Dialogue Turns & 593,156\\
    Average Turns per call & 27.95\\
    Average Tokens per Call & 544.16\\
    Average Tokens per Turn & 19.47\\
    \hline

    \end{tabular}
    \caption{Summary of the dataset regarding number of calls, human utterances (i.e., dialogue turns), and tokens.}
    \label{tab:dialogue_info}
\end{table}


\begin{table}
    \small
    \centering
    \begin{tabular}{l|lll}
        \hline
          Panel & Progress & Actions &  Dialogue Turns (\%) \\
         \hline
         0\textsuperscript{*} & 0\% & 17 &  313214(53\%)   \\
         \hline
         1 & 20\% & 39 &  43,095(7\%)   \\
         2 & 40\% & 4 &  135,52(2\%)   \\
         3 & 60\% & 4 &  166,068(28\%)   \\
         4 & 80\% & 20 &  57,227(10\%)   \\
         \hline
         Total & - & 80 & 593,156(100\%) \\
         \hline
         \multicolumn{4}{l}{*Panel 0: Authentication; finishing a call at this panel} \\
         \multicolumn{4}{l}{~~~means the call has failed.}
    \end{tabular}
    \caption{Summary of actions and dialogue turns per panels.}
    \label{tab:action_info}
    \vspace{-2mm}
\end{table}

\section{System Configurations}\label{app:sys_config}

The experiments in this paper including training and testing phases were done by two Computing Engines of the Google Cloud Platform; One including two ``NVIDIA T4 16 GB Memory'' GPUs and the other including a ``NVIDIA A100 40 GB Memory'' GPU.
``T4'' GPUs were used to train the MLM and GaLT models as well as other language transform approaches while the ``A100'' unit was used for GNN based approaches as they needed more memory.

\section{Models' Hyper-parameteres}\label{app:other_model_params}

Table~\ref{tab:params} lists the parameters and their values for training the proposed model.

\begin{table}
    \small
    \centering
    \begin{tabular}{l l}
         Parameter & Value (GaLT/MLM) \\
         \hline
         Epochs & 3/30 \\
         Batch Size & 256/512 \\
         Optimizer & AdamW/AdamW\\
         Max. Learning Rate & 5e-5/5e-5\\
         Learning Rate Policy & linear/linear\\
         Warmup steps & 250/250 \\
         \hline
         Max. Input Sequence Length & NA/128 \\
         Masking Probability & NA/15\%\\
         \hline
    \end{tabular}
    \caption{List of hyper-parameters the proposed models was trained on.}
    \label{tab:params}
\end{table}

\begin{figure}
    \centering
    \includegraphics[width=1.3\linewidth]{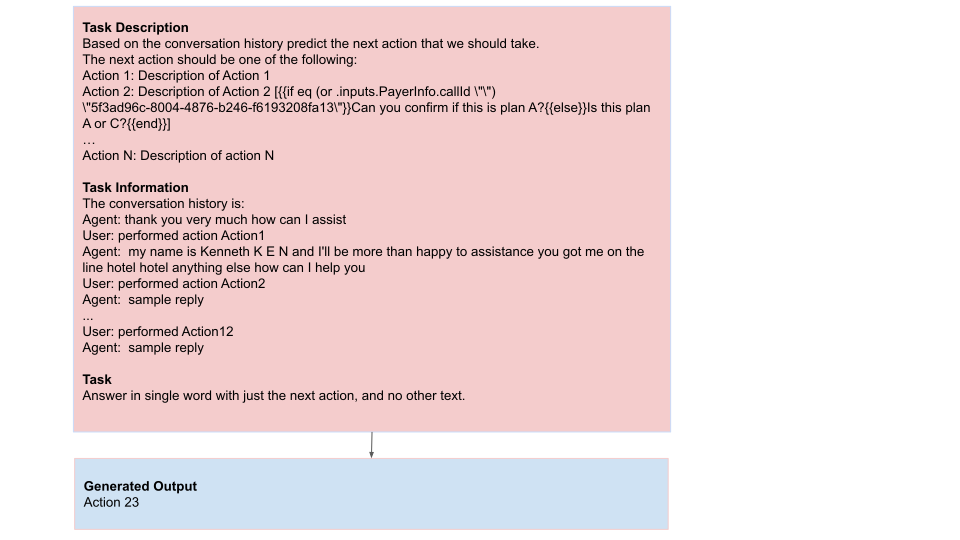}
    \caption{A sample snippet of the prompt input that is fed to the Llama2. Each Action has a name and a description that includes a Golang code on how the requirements of a next action is met. The prompt follows by the dialogue history and the task to generate response for. }
    \label{fig:propmt}
\end{figure}

\section{Prompting Llama2}\label{app:prompt}

To evaluate the performance of Prompt Engineering on Large Language Models, Llama~2 was chosen as it was one of the few available ones at the time of experiments that had the proper HIPAA compliance requirements in place which is a requirement for this healthcare dataset to stay PHI compliant~\footnote{\href{https://www.hhs.gov/answers/hipaa/what-is-phi/index.html}{https://www.hhs.gov/answers/hipaa/what-is-phi/index.html}}. The prompt included basic instructions on the task and the dialogue history between user (i.e., agent) and system (i.e., user). The model was evaluated on the same dataset and achieved an $F1$ score of 0.09. 
Figure~\ref{fig:propmt} shows a sample snippet of the prompt; it is customized for each request. The contextual information supplied in the prompt included basic instructions, the last actions of dialogue history (i.e., up to 10 turns), a list of next actions and their descriptions. To reduce prompt size and restrict the action search space, the prompt included only a subset of potential next actions. This list was determined by their observed co-occurrence in the dataset. All next actions were included if there were up to 10 co-occurring actions. For cases where there were more than 10, as many actions as required to add up to a cumulative sum of 50\% were added to the set.

\section{Data Size Effect} \label{app:data_effect}


Table~\ref{tab:data_effect} shows the effect of training size on the proposed models performance. It suggests having $60K$ training data is almost enough to train GaLT to perform close to its best.

\begin{table}
    \centering
    \begin{tabular}{l|ll}
         Train Size (\%) &   $F1_{Weighted}$ & $F1_{Macro}$ \\
         \hline
         5,930 (1\%) & 0.52 & 0.29\\
         11,860 (2\%) & 0.75 & 0.59\\
         59,300 (10\%) &  0.82 & 0.69\\
         296,500 (50\%) & 0.84 & 0.72 \\
         \hline
         593,156 (100\%) & 0.84 & 0.75 \\
    \end{tabular}
    \caption{Effect of training size on the proposed model, GaLT, performance.}
    \label{tab:data_effect}
\end{table}

\begin{figure*}
    \centering
    \begin{subfigure}[b]{0.45\textwidth}
        \centering
        \includegraphics[width=7cm]{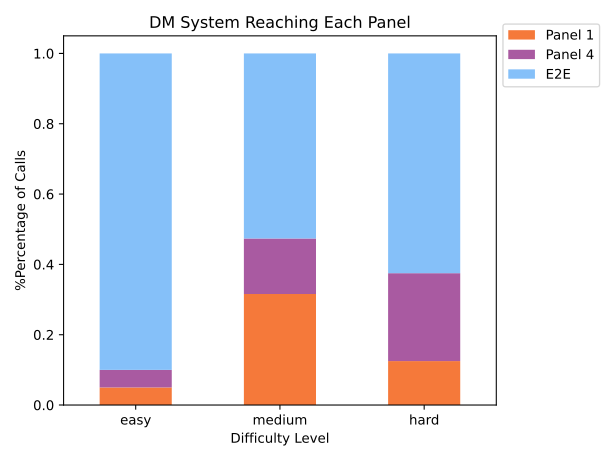}
        \label{fig:current_panel_num}
    \end{subfigure}
    \begin{subfigure}[b]{0.45\textwidth}
        \centering
        \includegraphics[width=7cm]{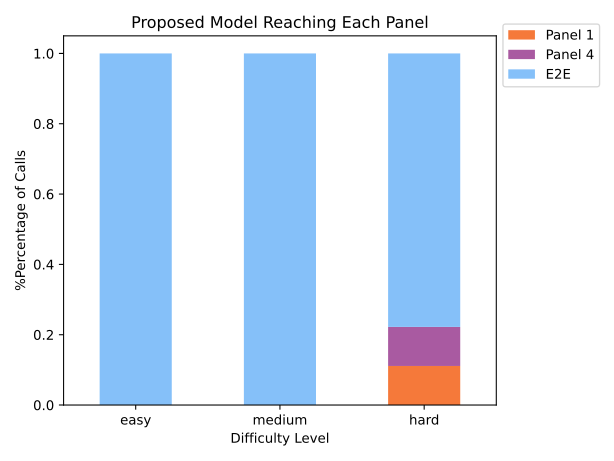}
        \label{fig:proposed_panel}
    \end{subfigure}
    \caption{Percentage of calls reaching to each panel for the DM system (i.e., left figure) and the proposed models (i.e., right figure). There were 4 panels and end of a call (i.e., \textit{E2E}) during a call and each model managed to finish only at panel 1, 4, or \textit{E2E} due to lower number of actions in panel 2 and 3.}
    \label{fig:panel}
\end{figure*}

\section{Human-centered Evaluation Setup}\label{app:human_centered_evaluation_setup}

A call is considered successful if the conversational AI system was able to prompt the recipient of the call to provide all information fields required for completion of the task.  
Therefore, the \textbf{number of fields} gathered by the system is a direct measure of call success.
The outbound call is structured into panels (i.e., \textbf{panel number}) which indicate how far into the conversation the system was able to navigate before call breakdown or completion. Therefore, both of these objective metrics indicate better performance the higher they are.
This paper computes both of these metrics as objective metrics.

Additionally, after finishing a call, the human agent is asked to rate the call from 1-5 (i.e., 1 being extremely dissatisfied and 5 extremely satisfied) using a Likert scale.
In addition to that, two additional human experts review both the objective and subjective assessment. 
Both the human agent and the reviewers answer an open-ended question of \textit{how they describe their experience with the system} at the end of the process.

To make sure the evaluation is not handled in error-free, lab settings but more similar to real-world settings, different levels of difficulties were defined and considered for each call (e.g., background noise or repeated expressions). Three difficulty levels were defined (i.e., hard, medium, and easy) and each level consisted of a minimum-maximum number of scenarios challenges described in Table~\ref{tab:difficulty-levels}.

\begin{table}
    \centering
    \begin{tabular}{l|lll}
         Scenario & \multicolumn{3}{c}{Level} \\
          & Easy & Medium & Hard \\
        \hline
        Agent & 0-1 & 2-3 & 4-5 \\
        Flow & 0-1 & 2-3 & 4-5\\
        \hline
    \end{tabular}
    \caption{Summary of how each difficulty level is made from agent and flow scenarios for calls. Human agents were assigned a difficulty level and could pick a number in the given range of the available scenarios to act out for their call. For each call difficulty level both agent and flow scenarios were picked from the same level.}
    \label{tab:difficulty-levels}
\end{table}

Two types of scenarios were defined; agent and flow scenarios. The agent scenarios (i.e., 5 conditions) are challenges regarding human users' performance during a call such as mumbling, background noise or repeated expressions. The flow scenarios (i.e., 6 conditions) are specific conditions and edge cases which increase the complexity of the conversation and the information required to be collected to complete the task.
For each level, both agent and flow scenarios are selected from the same difficulty level.

\section{Extended Results of Product-level Metrics}\label{app:extended_product_results}

\textit{E2E} metrics shows the proportion of calls a model can finish successfully. Put differently, reaching to panel 4 alone is not the desired goal but reaching to \textit{E2E} is the main goal of the call automation task. A comparison between the ``DM sytem'' and the proposed models in Figure~\ref{fig:panel} shows that the proposed models are perfect (i.e., 100\% \textit{E2E} reach) in \textit{easy} and \textit{medium} difficulties. 
However, the DM system was only able to finish 90\% of the easy and 70\% of medium calls.
The proposed models also managed to finish $78\%$ of calls successfully with \textit{hard} difficulty whereas the DM system were able to finish $62\%$ of calls.

\section{Extended Results of Subjective Human Evaluation}\label{app:extended_subjective_results}

\begin{figure}
    \centering
    \includegraphics[width=1\linewidth]{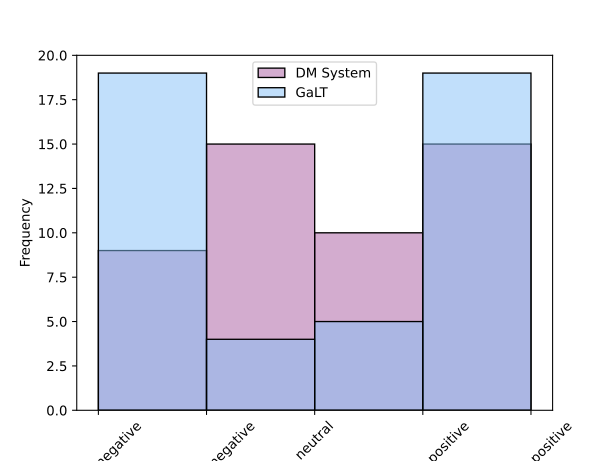}
    \caption{Distribution of human evaluation ratings for calls managed by the DM system and proposed model. }
    \label{fig:rating_total}
\end{figure}

Figure~\ref{fig:rating_total} shows the distribution of the ratings regarding each model. The ''DM system`` received negative ratings (i.e., strongly negative) twice as much as the proposed model.

\begin{figure}
    \centering
    \includegraphics[width=1\linewidth]{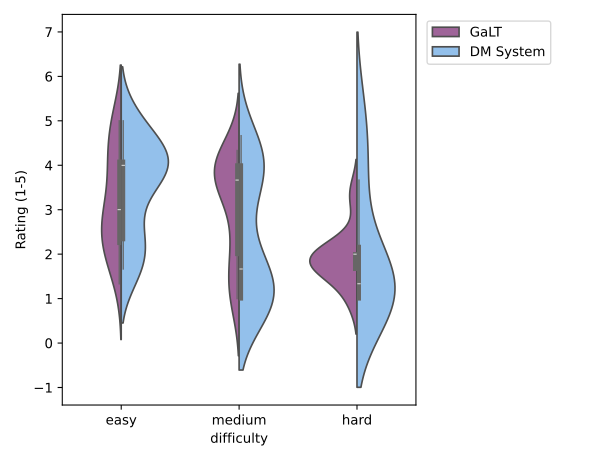}
    \caption{Violin plot of distribution for human evaluation rating regarding difficulty levels.}
    \label{fig:rating_levels}
\end{figure}

Figure~\ref{fig:rating_levels} shows the distribution of the models regarding the human ratings using a violin plot. The DM system performed better regarding human assessment across \textit{easy} difficulty ($M_{Current}=3.48 > M_{Proposed} = 3.2;  STD_{Current}=1.08 < STD_{proposed}=1.11$). 
The DM system also performed slightly better for calls with \textit{hard} difficulty on average but with a much larger standard deviation ($M_{Current}=2.00 > M_{Proposed} = 1.93; STD_{Current}=1.41 >  STD_{proposed}=0.58$). 
Higher standard variation for \textit{hard} difficulty indicates that the proposed models will generate less unexpected actions or outputs.
Moreover, the proposed models outperformed the DM system across \textit{medium} difficulty ($M_{Current}=2.37 < M_{Proposed} = 3.07; STD_{Current}=1.41 >  STD_{proposed}=1.14$).

However, t-test statistics of human ratings across different difficulties as well as all levels combined were \underline{not} significant ($p>0.1$). These findings suggests, a careful consideration when interpreting these results and perhaps the need for a larger sample to compare ratings for both models.

\end{document}